\documentclass[10pt,twocolumn,letterpaper]{article}

\usepackage{pgf}
\usepackage{iccv}
\usepackage{times}
\usepackage{epsfig}
\usepackage{graphicx}
\usepackage{amsmath}
\usepackage{amssymb}
\usepackage[lined,boxruled,linesnumbered]{algorithm2e}
\usepackage{graphicx}
\usepackage{color}
\usepackage{dsfont}
\usepackage{bbm}
\usepackage{subcaption}
\usepackage{multirow}
\usepackage{rotating}
\usepackage{enumerate}
\usepackage{enumitem}

\def\w{\boldsymbol{w}}

\def\x{\boldsymbol{x}}

\def\bpsi{\boldsymbol{\psi}}
\def\bphi{\boldsymbol{\phi}}
\def\ourmethod{Budget~}

\usepackage[pagebackref=true,breaklinks=true,letterpaper=true,colorlinks,bookmarks=false]{hyperref}

\iccvfinalcopy % *** Uncomment this line for the final submission

\def\iccvPaperID{593} % *** Enter the ICCV Paper ID here

\ificcvfinal\fi
\begin{document}

%%%%%%%%% TITLE
\title{Approximate Policy Iteration for Budgeted Semantic Video Segmentation}

\author{Behrooz Mahasseni,  Sinisa Todorovic, and  Alan Fern\\
Oregon State University\\
 Corvallis, OR 97331, USA\\
{\tt\small mahasseb@eecs.oregonstate.edu, sinisa@eecs.oregonstate.edu, afern@eecs.oregonstate.edu}
% For a paper whose authors are all at the same institution,
% omit the following lines up until the closing ``}''.
% Additional authors and addresses can be added with ``\and'',
% just like the second author.
% To save space, use either the email address or home page, not both
}

\maketitle
%\thispagestyle{empty}

%%%%%%%%% ABSTRACT
%%%%%%%%%%%%%%%%%%%%%%%%%%%%%%%%%%%%%%%%%%%%%%%%%%%%%%%%%%%%%%%%%%%%%%%%%%%%%%%%%%%%%%%%%%%%%%%%%%%%%%%%%%%%%%%%%%%%%%%%%%%%%%%%%%%%%%%%%%%%%%
\begin{abstract}
This paper formulates and presents a solution to the new problem of budgeted semantic video segmentation. Given a video, the goal is to accurately assign a semantic class label to every pixel in the video within a specified time budget. Typical approaches to such labeling problems, such as Conditional Random Fields (CRFs), focus on maximizing accuracy but do not provide a principled method for satisfying a time budget. For video data, the time required by CRF and related methods is often dominated by the time to compute low-level descriptors of supervoxels across the video. Our key contribution is the new budgeted inference framework for CRF models that intelligently selects the most useful subsets of descriptors to run on subsets of supervoxels within the time budget. The objective is to maintain an accuracy as close as possible to the CRF model with no time bound, while remaining within the time budget. Our second contribution is the algorithm for learning a policy for the sparse selection of supervoxels and their descriptors for budgeted CRF inference. This learning algorithm is derived by casting our problem in the framework of Markov Decision Processes, and then instantiating a state-of-the-art policy learning algorithm known as Classification-Based Approximate Policy Iteration. Our experiments on multiple video datasets show that our learning approach and framework is able to significantly reduce computation time, and maintain competitive accuracy under varying budgets.
\end{abstract}
%%%%%%%%% INTRODUCTION
%%%%%%%%%%%%%%%%%%%%%%%%%%%%%%%%%%%%%%%%%%%%%%%%%%%%%%%%%%%%%%%%%%%%%%%%%%%%%%%%%%%%%%%%%%%%%%%%%%%%%%%%%%%%%%%%%%%%%%%%%%%%%%%%%%%%%%%%%%%%%%
\vspace{-2em}
\section{Introduction} \label{sec:introduction}

{\bf Motivation:} The design of vision approaches is typically informed by a trade-off between efficiency and accuracy. Good computational efficiency is usually achieved by taking a number of heuristic pre- and post-processing steps and integrating them with the main approach. For example, vision practitioners heuristically limit the types of features to be extracted (e.g., low-cost ones), as well as locations and scales in images and video from which they are extracted. While these steps have been satisfactory for small-scale problems, their heuristic nature makes an adaptation of existing systems to settings with stringent runtime requirements very difficult.

{\bf Our goal} is to formulate a principled framework for optimally adapting a vision system to varying time budgets imposed by particular application settings, so as to maximize overall performance for any budget and maintain an accuracy as close as possible to the system's performance with no time bound. In this paper, we focus our presentation of theory and experiments in the context of semantic video segmentation. Nevertheless, it is worth noting that our framework is based on fairly non-restrictive assumptions, and thus is suitable for many other computer vision problems, beyond the scope of this paper.

{\bf The Focus Vision Problem:} Given a video, our goal is to assign a class label to every pixel from the set of semantic classes seen in training, under any  time budget. We call this new problem \emph{budgeted semantic video segmentation}. For example, in a video of a street, we want to efficiently segment spatiotemporal subvolumes occupied by cars, pedestrians, and buildings in less time than the user-specified bound.  This is an important problem with a wide range of applications (e.g., driverless cars, sports video analytics) which require highly accurate and timely estimates of space-time extents of objects in the scene.

{\bf The key idea:} We assume that a given approach to semantic video segmentation specifies an inference procedure (e.g., loopy belief propagation, graph-cut) which takes input features and outputs an inference result. Rather than pixels, most existing approaches, first, label supervoxels, obtained from an unsupervised (low-level) video partitioning, and then transfer these labels to the corresponding pixels. Thus, a typical input consists of supervoxels and their {\em descriptors}, where the descriptors may be computed locally at every supervoxel, between pairs of neighboring supervoxels, and globally across the entire video. Computing all descriptors for all supevoxels is costly. Thus, a sparse section of supervoxels and choosing their most useful and least costly descriptors is our key problem. Note that we do not modify the inference procedure. Rather, we adapt the descriptor computation step to suit budgeted settings of a broad family of approaches. Consequently, we do not seek to improve prediction accuracy, but rather maintain the existing level of accuracy while reducing runtime.

%
%%%%%%%%%%%%%%%%%%%%%%%%%
\begin{figure}
\centering
\includegraphics[width=\columnwidth]{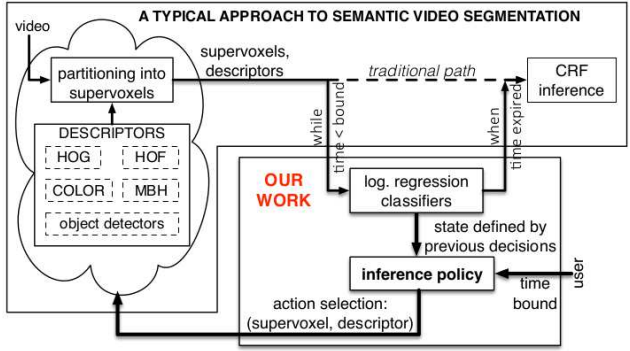}
\caption{We formalize a sequential inference policy aimed at adopting a fairly general family of approaches to the problem of budgeted semantic video segmentation. Our focus domain is a holistic CRF-based inference, but other approaches to semantic video segmentation could be considered. Given an unsupervised video partitioning into supervoxels, a set of feature extractors, and a user-specified time budget, our policy sequentially selects the best pair (supevoxel, feature) toward maximizing performance of the CRF inference until the time expires.}
\label{fig:BlockDiagram}
\vspace{-1em}
\end{figure}

{\bf Our framework} is illustrated in Fig.~\ref{fig:BlockDiagram}. We assume access to a Conditional  Random Field (CRF) and associated inference procedure---one of the most popular formulations of semantic video segmentation---and design a {\em sequential} inference policy that interacts with the given  approach. Given an unsupervised video partitioning into supervoxels and a user-specified time budget, our policy is presented with a sequence of candidate supervoxels until time expires and must decide for each one which new descriptor to run for it, if any, with the goal of maximizing performance of the CRF inference. The sequential selection should take into account both the immediate and long-term value of the decisions toward overall inference accuracy.

We formalize such sequential decision making in the framework of Markov Decision Processes (MDPs), where, for a given state, the policy selects the highest value action among possible actions. The policy state is defined by the previous selections of descriptors for supervoxels, which is conveniently represented via special policy features. The policy actions include running a descriptor for the currently presented supervoxel or {\sc Finished}, which specifies that no further descriptors should be computed for the current supervoxel. The policy is defined as a {\em linear} ranking function for balancing efficiency and expressiveness, such that its execution consumes resources negligibly, and that it captures sufficient information about the current policy state to support good decisions. For training the policy, we use the state-of-the-art policy learning approach of Classification-Based Approximate Policy Iteration (CAPI), which is able to leverage state-of-the-art classification learning techniques for policy optimization.

\section{Related Work and Our Contributions}\label{sec:priorwork}

%%% Video labeling prior work
Semantic video segmentation is mostly formulated as a graphical-model based labeling of supervoxels in the video  \cite{chen2010propagating,budvytis2010label, brostow2008segmentation,miksik2013efficient, chen2011temporally,liumulti,taylor2013semantic,wojek2008dynamic}. For example, graphical models were used for: i) Propagating manual annotations of supervoxels of the first few frames to other supervoxels in the video \cite{miksik2013efficient, chen2010propagating, budvytis2010label}, or ii) Supervoxel labeling based on week supervision in training \cite{Liu_2014_CVPR}. The accuracy of such labeling can be improved by CRF-based reasoning about 3D relations \cite{kundu2014joint} or context \cite{liumulti} among object classes in the scene. Therefore, it seems reasonable that we develop our framework for an existing CRF-based labeling of supervoxels. None of these methods explicitly studied their runtime efficiency, except for a few empirical results of sensitivity to the total number of supervoxels used \cite{jain2013coarse}.

Prior work has considered the issue of how to reduce inference runtime by specifying efficient approximations to the original inference \cite{zhang2012efficient,krahenbuhl2012efficient, a22001fast}. Their work is not related to ours, since we keep the given inference procedure intact. A few approaches have addressed cost-sensitive inference for activity recognition \cite{Mohamed_ECCV12, Amer_ICCV13}, image classification \cite{Efficiency_CVPR13, AnytimeRecognition_CVPR14}, and object detection \cite{Wu_ICCV2013}. Our fundamental difference is that these methods precompute all features and then adapt the very inference procedure such that it uses only an optimal subset of the features to meet the time budget. In contrast, we do not compute any features before our inference policy makes the decision to do so.

Closely related work improves runtime efficiency by reducing the costs of feature extraction \cite{Weiss_2013_ICCV, NIPS2013_5142, roig_active_2013, liu2003multiclass, grubb2013speedmachines}.  For example, CRF-based semantic scene labeling in images is made more efficient by computing only a small subset of unary potentials for the CRF, and efficiently predicting the missing potentials from neighbors \cite{roig_active_2013}. This approach can be viewed as a special case of our framework, since they only select superpixels and then use {\em all} features for computing the unary potentials, whereas we would select both superpixels and feature types in their domain. In \cite{NIPS2013_5142}, a budget constrained reinforcement learning is used to select optimal features for tracking human poses in relatively simple videos. Since they extract a chosen feature over the entire video, this approach can be viewed as another special case of our framework, because we would additionally make the decision about where to extract the feature in the video. This is a crucial difference for large videos, where computing even a single feature/descriptor across an entire video may exceed the budget constraint. Additional differences from these two approaches are explained in Sec.~\ref{sec:problem-statement}.  We efficiently compute low-level descriptors for a selected subset of supervoxels in addition to the object level potentials in \cite{liu2003multiclass}. Unlike \cite{grubb2013speedmachines} we do not change the inference module, which makes it possible to augment any state of the art with our approach. It is not clear how to extend \cite{grubb2013speedmachines} to use higher order potentials. More importantly our approach is able to exploit the high correlation among the features of neighboring supervoxels to avoid the extra feature extraction cost if it does not help the final inference.

{\bf Contributions:} 1) First formalization of the budgeted semantic video segmentation problem. 2) Design of the first learning algorithm that can tune inference policies for varying time budgets. 3) Specification and evaluation of the inference policy representation as a linear function that supports learning.

%%%%%%%%%%%%%%%%%%%%%%%%%%%%%%%%%%%%%%%%%
\section{Budgeted Semantic Video Segmentation} \label{sec:problem-statement}

Our framework takes as input a time budget $B$ and video that has been partitioned into a set of supervoxels $V$. The goal is to accurately assign a label to each supervoxel in $V$ in time less than $B$. Below, we first present a common CRF formulation for semantic video segmentation, which our framework is based on. Next, we formulate our framework for bounding the CRF inference time.

%%%%%%%%%%%%%%%%
\subsection{A Common CRF Formulation}\label{sec:CRF}
We consider a standard pairwise CRF model, which specifies the following score for any labeling $\{y_i\}$ of supervoxels $i \in V$ and their spatiotemporal neighborhood relationships $(i,j)\in E \subset V\times V$:
\setlength{\arraycolsep}{0pt}
\begin{eqnarray} \label{eq:CRF}
\sum_{i \in V}  \w_u\cdot \bpsi_u(\x_i,y_i) {+}
 \sum_{(i, j) \in E} \w_p\cdot \psi_p(\x_i,\x_j,y_i,y_j),
\end{eqnarray}
where $\w_u\cdot \bpsi_u(\x_i,y_i)$ denotes the unary potential specified in terms of  an $n_u$-dimensional \emph{unary feature vector} $\bpsi_u(\x_i,y_i)$ estimated for observation $\x_i$ at supervoxel $i$ when $i$ is  assigned label $y_i$ from the set of labels $\mathcal{L}$, and the corresponding weight vector $\w_u$. Also, we have that $\w_p\cdot \psi_p(\x_i,\x_j,y_i,y_j)$ assigns pairwise ``compatibility" scores of assigning labels $y_i$ to $i$ and $y_j$ to neighboring supervoxel $j$, where  $\bpsi_p(\x_i,y_i,\x_j,y_j)$ is an $n_p$-dimensional \emph{pairwise feature vector} and $\w_p$ as the associated weight vector. We specify $n_u = L = |\mathcal{L}|$ and  $n_p=L^2$, but more general specifications are also possible.

We consider a common form of unary features defined in terms of a probabilistic multi-class classifier. In particular, we will learn a probabilistic classifier $H$ that returns an $L$-dimensional probability vector, such that $H(i)$ is a predicted distribution over all labels $y_i\in\mathcal{L}$ for supervoxel $i$. The input to $H$, used as a basis for prediction, is the concatenation of multiple descriptors of supervoxel $i$. In this paper, we use logistic regression for $H$. Given $H$, the unary feature vector $\bpsi_u(\x_i,y_i)$ is constructed in the standard way by setting all elements of $\bpsi_u(\x_i,y_i)$ to zero, except for the element corresponding to label $y_i$ which is set to the probability $H(i)$ of predicting $y_i$ for $i$. Thus, the cost of $\bpsi_u(\x_i,y_i)$ is dominated by the cost of computing $H(i)$.

The pairwise features is specified in the standard way as the difference between descriptors of the pair of neighboring supervoxels. Thus, the pairwise ``compatibility" score is defined to increase as this difference becomes smaller for the supervoxels with the same label, and to decrease for small descriptor differences when the pair of supervoxels have different labels. Non-overlapping features are replaced by the expected features conditioned on the label.

{\bf CRF inference} involves two main steps. First, unary and pairwise potentials are computed for all combinations of supervoxels and labels. Second, a standard approximate CRF inference procedure is applied, such as belief propagation or $\alpha$-expansion (used in our experiments), which returns a high scoring (ideally optimal) label assignment. The overall computational cost is, thus, the total time for computing the potentials and running the inference procedure.

{\bf Learning the CRF and $H$.} Given labeled training data, we learn our CRF model, $\mathcal{M}$, using the unary and pairwise feature vectors computed over all training supervoxels. This is done by computing all descriptors for all supervoxels in the training data, and then using a standard CRF library to obtain $\mathcal{M}$ and the associated logistic regression $H$.

Recall, that the main idea of our framework is to reduce time by only computing a subset of descriptors for some supervoxels. Thus, in order to produce the unary feature vector, $H$ must be able to make label predictions for a supervoxel using any subset of its descriptors. One approach to obtaining such predictions would be to train $H$ on all descriptors, and then use one of a number of common strategies for making predictions with ``missing information" (e.g., replacing the missing descriptor values with zero or expected values).

Rather, we take a more brute force approach, with the advantage of not requiring any method for handling missing descriptors. Instead of just training a single classifier based on all descriptors, we train a {\em collection of classifiers}, one classifier for each possible subset of descriptors. $H$ is then represented by this collection of classifiers. That is, each logistic regression is trained by removing all descriptor information from the training set other than its assigned descriptor subset. When $H$ is asked to make a prediction for a supervoxel during budgeted inference, it uses the classifier corresponding to the set of descriptors that have been run for that supervoxel. Given that classifier training is very fast, this will often be practical even with thousands of possible subsets. This is indeed the case in our experiments as described in Sec.~\ref{sec:results}.

%%%%%%%%%%%%%%%%%%%%
\subsection{Budgeted Unary Feature Computation}\label{sec:budgeted}

Computing the unary features $\bpsi_u$ typically dominates overall computation time, since this involves computing a number of low-level descriptors over all supervoxels. The pairwise  features $\bpsi_p$ are much cheaper, in comparison, since they are generally based on comparing descriptors already computed for $\bpsi_u$. For this reason, our framework focuses on bounding the time of computing $\bpsi_u$, and we will let $B$ denote this time bound for the remainder of this paper. The overall CRF inference  will typically be a small constant larger than $B$, when $B$ is non-trivial.

Our hypothesis is that similar accuracies in inference can be achieved with less cost by intelligently selecting for each supervoxel a sparse set of descriptors to compute, including the empty set, which are then used by $H$ to generate $\bpsi_u$. Below, we describe how to make these decisions.

Alg.~\ref{alg:inference} presents our iterative approach to selecting descriptor subsets for time bounded inference. Throughout the iterations, we maintain a set of \emph{candidate supervoxels}, $\mathcal{C}$, which are currently being considered for descriptor computation. $\mathcal{C}$ is initialized to a small random subset of $V$. We also maintain a set of \emph{finished supervoxels}, $\mathcal{F}$, which will no longer be considered for descriptor computation. Each iteration consists of two steps. The first step calls the function $\mbox{Select}(\mathcal{C})$, which returns a supervoxel $i \in \mathcal{C}$ to be considered next. As described in Sec.~\ref{sec:results}, we consider two versions of $\mbox{Select}(\mathcal{C})$: a) Random selection, and b) Priority-based selection. The second step applies \emph{policy} $\pi$ for selecting either a new descriptor for $i$ to compute, or mark $i$ as being finished. Specifically, $\pi(i)$ returns an \emph{action} for $i$ that is either a descriptor index or {\sc Finished}. In the latter case, $i$ is moved from $\mathcal{C}$ to $\mathcal{F}$, and all neighbors of $i$ that are not already in $\mathcal{F}$ are added to $\mathcal{C}$. The iterations continue until the runtime reaches $B$.  When $B$ is reached, the CRF unary features $\bpsi_u(\x_i,y_i)$ are computed for all supervoxels $i$ where at least one descriptor has been computed. For a supervoxel $i$, where no descriptors are computed, $\bpsi_u(\x_i,y_i)$ is estimated based on the available unary features $\{\bpsi_u(\x_j,y_j)\}$ of $i$'s neighbors, i.e. $\bpsi_u(\x_i,y_i)$ is set to a weighted sum of $\{\bpsi_u(\x_j,y_j)\}$, where the weights are the inverse Euclidean distances between the centroids of $i$ and the neighbors.

\begin{algorithm}
{\small
 \KwIn{Supervoxels $V$, Policy $\pi$, Classifier $H$, Budget $B$}
 \KwOut{Labeling of supervoxels}
  $\mathcal{C} \leftarrow \mbox{small random subset of } V$;
  $\mathcal{F} \leftarrow \emptyset$\;
  \While {runtime $< B$} {
    $i \leftarrow \mbox{Select}(\mathcal{C})$  \% select a supervoxel (see text)\;
    $a \leftarrow \pi(i)$   \% apply policy on $i$ \;
    \If {$a == \mbox{\sc Finished}$} {
        $\mathcal{F} = \mathcal{F} + \{i\}$ \;
        $\mathcal{C} = \mathcal{C} + \mbox{Neighbors}(i) - \mathcal{F}$  \% new candidates
    }
    \Else {
        Compute the descriptor specified by $a$ for $i$
    }
  }
  Interpolate unary features for supervoxels $i$ with no computed descriptors (see text)\\
  Apply CRF inference
 \caption{Our cost-sensitive inference}
  \label{alg:inference}
  }
\end{algorithm}

The key element in the above framework is the policy $\pi$. Given a supervoxel $i$, $\pi$ must weigh the cost of computing a new descriptor of $i$ versus the potential improvement in accuracy of the CRF inference. It is critical that $\pi$ makes these decisions efficiently, otherwise the cost of evaluating $\pi$ would negate any potential reduction in descriptor computation that it provides. As our key contributions, in the following Sec.~\ref{sec:learn},  we specify a suitable representation and learning algorithm for such a $\pi$.

\section{Policy Representation and Learning}\label{sec:learn}

This section first describes the linear representation we use for policies, and then formulates our policy learning algorithm within the MDP framework.

\subsection{Policy Representation} \label{sec:representation}

At each iteration of our budgeted inference, $\pi$ is shown a supervoxel $i$, and asked to decide whether or not to run a new descriptor for $i$ and if so which one. We call these choices actions. $\pi$ selects an action $a$ based on the information available at the time, which we call the \emph{inference state} $s$, $a=\pi(s)$. An inference state is a tuple $s=(i,b,\mathcal{C},\mathcal{F},\mathcal{D})$, where $i$ is the supervoxel currently being considered, $b$ is the remaining budget, and $\mathcal{C}$ and $\mathcal{F}$ are the sets of candidate and finished supervoxels as described in Sec.~\ref{sec:budgeted}. Finally, $\mathcal{D}$ is the set of descriptor outputs that have been produced so far for supervoxels in $\mathcal{C}$ and $\mathcal{F}$.

Since $\pi$ is called many times during inference, it is critical that the time required to select an action be significantly smaller than the time required to run descriptors. To support efficiency we represent $\pi$ as a linear function that ranks the possible actions at an inference state $s$ based on an easy to compute vector of $\pi$-features $\bphi(s)$.

{\bf Policy Features.} The $\pi$-features have three subvectors $\bphi(s)=[\bphi_1(s),\bphi_2(s),\bphi_3(s)]$ that capture different aspects of the inference state $s$. $\bphi_1(s)$ is a binary vector that indicates which descriptors have already been run for $i$. $\bphi_1(s)$ allows the policy to learn the value of taking certain actions, given various combinations of computed descriptors characterizing $s$. $\bphi_2(s)$ is a vector that is equal to a weighted average of the unary potential features $\{\bpsi_u\}$ of ``finished" neighbors of $i$ that are in $\mathcal{F}$. Recall that $\bpsi_u$ corresponds to probability distributions over class labels. Thus, $\bphi_2(s)$ allows the policy to base its decisions on the confidence of neighboring supervoxels about the various semantic labels. For example, the policy can learn that if all neighbors are very confident about a particular label then it is not worth computing further descriptors for $i$. Finally, $\bphi_3(s)$ is the standard shape-context descriptor capturing the spatiotemporal layout of finished supervoxels in $\mathcal{F}$ around $i$. $\bphi_3(s)$ is computed by binning the space-time neighborhood of $i$ (all supervoxels that touch $i$) into 8 bins ($\{up, down, left, right\} \times \{before, after\}$), and counting the finished supervoxels that fall in each bin. $\bphi_3(s)$ allows the policy to base its decisions in part on the density of surrounding finished supervoxels.

Given the $\pi$ features for an inference state $s$, $\bphi(s)$, our policy is a linear ranking function over policy actions $a$, represented in terms of a weight vector $\w_a$ for each action.
\begin{equation}
\pi(s)=\arg\max_a ~\w_a\cdot \bphi(s)
\end{equation}
Thus, training the policy entails training the weights $\w_a$ so as to make the best possible decisions.

\subsection{Policy Learning}\label{sec:pi-learning}
It is important to note that ground-truth annotations of our training videos do not directly provide ground-truth decisions that should be made by $\pi$.  The supervised training data for $\pi$ would need to label inference states by best policy actions. Since this information is not available in training, $\pi$ cannot be learned via pure supervised learning. The training data does, however, provide the means for evaluating the quality of any policy. In particular, given any $\pi$ and budget $B$, we can run time-bounded inference on each training video using $\pi$, as summarized in Alg.~\ref{alg:inference}, and then measure the prediction accuracy of the CRF relative to the available ground truth. In practice, large numbers of such policy evaluations can be run quickly by precomputing all descriptors for all supervoxels across the training videos. This allows for the budgeted inference process to be ``simulated" on the training data without requiring descriptors to be recomputed for each policy evaluation. The question then is how to use this fast policy evaluation on the training data in order to learn an effective policy?

Policy learning is complicated by the fact that the policy is inherently solving a sequential decision making problem, where each decision may have long-term impacts on the overall solution accuracy. Optimizing the long-term value of policies is challenging due to the fact that each inference process will involve many policy decisions and assigning relative credit to those decisions toward the overall accuracy is non-trivial. Such sequential decision making problems are naturally formalized in the MDP framework \cite{MDP_1994}.

{\bf MDP Formulation:} An MDP specifies a set of states $S$, a set of actions $A$ that can be taken by a policy, and a transition function $T$, which describes how the state of the system changes when actions are taken. In addition, an MDP specifies a reward function, which evaluates the relative goodness of various system states. In our time-bounded inference application, the states correspond to inference states $(i,b,\mathcal{C},\mathcal{F},\mathcal{D})$ as described in Sec.~\ref{sec:representation}. The actions $A$ correspond to policy actions of either selecting to run a descriptor for the current supervoxel $i$, or returning {\sc Finished}, which indicates that we are finished computing descriptors for $i$.

The transition function describes how an action $a$ changes a state $s=(i,b,\mathcal{C},\mathcal{F},\mathcal{D})$. If the action $a$ is to run a descriptor, then the new state is equal to $s'=(i',b',\mathcal{C},\mathcal{F},\mathcal{D}')$, where $\mathcal{D}'$ updates $\mathcal{D}$ with the newly computed descriptor information for $i$. In addition, the new budget $b'$ will be equal to $b$ minus the cost of $a$, and $i'\in \mathcal{C}$ will be the supervoxel selected next for processing. When the action is {\sc Finished}, the new state is $s'=(i',b,\mathcal{C}',\mathcal{F}',\mathcal{D})$, where $\mathcal{C}'$ and $\mathcal{F}'$ are updated as described in Sec.~\ref{sec:budgeted}, and $i'$ is the newly selected supervoxel. Note that when $b=0$, no further actions are allowed. Importantly, the reward function is zero for all states, except for final states with $b=0$, where the reward is equal to the accuracy achieved by the CRF inference using the selected descriptors run by the $\pi$.

Given the above MDP formulation, the problem of optimizing $\pi$ to maximize expected long-term reward in the MDP is identical to finding a policy that maximizes inference accuracy within our budgeted inference framework. Thus, in principle, any policy learning algorithm from the MDP literature could be employed for our problems. Prior work \cite{NIPS2013_5142} used reinforcement learning (RL) for learning an approximate Q-function. In that work, the Q-function $Q(s,a)$ of an MDP gives the expected future reward of being in state $s$ and taking action $a$. Given $Q(s,a)$ the policy is defined to select the action with largest Q-value. Unfortunately, the Q-function can be extremely complicated to represent for problems that involve long sequences of decisions. In that prior work, the number of decisions was bounded by the number of descriptors, which is quite small. Rather, in this paper, the number of decisions is related to the number of supervoxels, which is substantially larger. Our early experiments showed that standard approaches for learning Q-functions, represented using our $\pi$-features, were ineffective for the problem scales we address here.

{\bf Classication-Based Approximate Policy Iteration (CAPI):} Our approach is motivated by the fact that we do not actually need to learn the Q-function, but only learn to rank good actions above bad actions, e.g., using the linear policy representation described in Sec.~\ref{sec:representation}. %Fortunately, it is often the case that such ranking functions can be significantly simpler to represent than the full Q-function, since they need only represent the decision boundary between good and bad actions.
This suggests considering approaches that directly learn the decision boundary between good and bad actions. Classification-Based Approximate Policy Iteration (CAPI) is one such state-of-the-art technique that we follow here.

CAPI was originally proposed by \cite{Fern03,Lagoudakis03}. It has demonstrated a number of empirical successes, and has been the subject of theoretical analysis providing various performance guarantees \cite{Fern06,Dimitrakakis08,Lazaric10}. A key distinction of CAPI is that it is able to leverage state-of-the-art learning algorithms for classification and ranking (e.g. SVMs).

CAPI is conceptually simple. Given an initial policy $\pi_0$, which is random in our experiments, CAPI iteratively applies an \emph{approximate policy improvement operator} $\mathcal{PI}$, which takes an input policy $\pi$ and returns an (approximately) improved policy $\mathcal{PI}[\pi]$. Thus, the CAPI algorithm produces a sequence of improving policies $\pi_{t+1} = \mathcal{PI}(\pi_t)$ and terminates when no further improvement is observed or a training time bound is reached. Recall that for our linear policy representation this will correspond to a sequence of weight vectors. It remains to describe the approximate policy improvement operator $\mathcal{PI}$.

Given a current policy $\pi$, CAPI computes the improved policy $\mathcal{PI}(\pi)$ using a two step process: {\bf Step 1 -- Training Set Generation.} Create a training set of state-action pairs $\mbox{Trn}=\{(s_i,a_i)\}$ such that $a_i$ is an ``improved" action for $s_i$, i.e. better or at least as good as the current action $\pi(s_i)$. {\bf Step 2 -- Classifier Learning.} Apply a classifier learner to $\mbox{Trn}$ to obtain a policy $\pi'$ that achieves high accuracy in selecting the improved actions. In Step 2 we use a multi-class SVM classifier to learn the weights of $\pi$ based on the training set generated by Step 1.

Step 1 requires selecting a set of states for the training set and then computing the improved actions. In our experiments, we have found that an effective and simple approach is to run the current policy $\pi$ on the training videos, and to let the training states correspond to all inference states encountered during inference using $\pi$ across all training videos. For each such inference state $s_i$ we must now compute a label $a_i$ that corresponds to an improved action relative to the action selected by $\pi$. 

\begin{figure}
  \centering
    \includegraphics[width=\columnwidth]{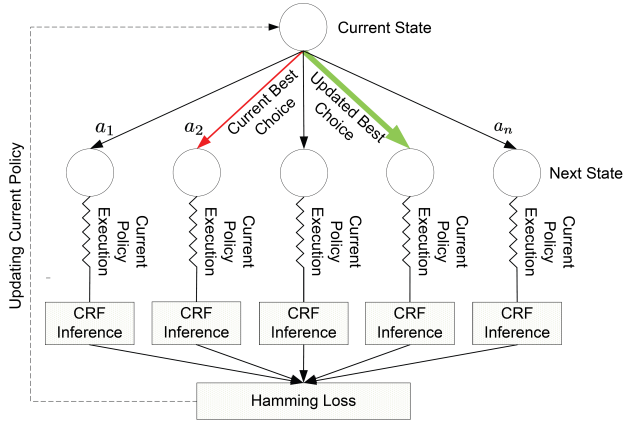}
  \caption{Monto Carlo simulation of policy rollout. All possible actions of a current state are evaluated by running the current policy $\pi$ starting from the next states corresponding to the actions being taken, until the time budget is reached. The CRF inference is then applied to compute the Hamming loss. This is used to improve the current policy.}
    \label{fig:policy_rollout}
    \vspace{-2em}
\end{figure}

In order to compute an improved action, CAPI uses the Monte Carlo simulation technique of \emph{policy rollout} \cite{Tesauro96}. Fig.~\ref{fig:policy_rollout} illustrates the main components of policy rollout. Simply stated, policy rollout computes a score for each action $a$ at a state $s_i$ that is equal to the accuracy achieved by the final CRF inference after taking action $a$ in $s_i$ and then taking actions according to $\pi$ until the budget is zero. The action leading to the highest accuracy is then selected. That is, rollout considers all one-step action departures from the current policy $\pi$ at $s_i$ and selects the action that resulted in the highest final CRF accuracy. Note that the transition function for actions may be stochastic, e.g., when the selection of the next supervoxel is implemented by random selection. In these cases, policy rollout runs multiple simulations for each action and the average accuracies across simulations are used to score actions. 

For deterministic transition functions, policy rollout is guaranteed to return an improved action if the current policy is not optimal in $s_i$. For stochastic transitions, in the limit of infinite simulations, the action returned by rollout is guaranteed to be an improved action. A polynomial sample complexity bound exists on the quality of the action returned by rollout compared to that of $\pi$ \cite{Fern06}. Since our classifier is linear the learning time is linear in sample size. Note that while the training process may require significant time, the end result is a single policy $\pi$ from the last iteration of CAPI that can be used efficiently at test time for time-bounded inference.

%%%%%%%%%%%%%%
\section{Results} \label{sec:results}

Our goal is to empirically support the claim that we can optimally adapt a given method for semantic video segmentation to varying time budgets, such that it yields satisfactory performance for any budget, and maintains an accuracy as close as possible to its performance for no time bound. As the given method was originally designed to perform best without time constraints, it is important to note that our performance is inherently upper-bounded by the method's accuracy for an infinite time budget. Therefore, our evaluation differs from much work in computer vision, where the focus is on demonstrating improvements in accuracy. 

\textbf{Datasets.} For evaluation, we use three benchmark datasets: 1) CamVid \cite{brostow2008segmentation}, 2) MPIScene \cite{wojek2010monocular}, and  3) SUNY Buffalo-Xiph.org 24-class \cite{chen2010propagating}. CamVid  consists of 5 videos with an average length of 5000 frames. The videos are captured with a moving camera recording road scenes. Following prior work, we focus on the 11 most common object class labels and use the standard split of training and test frames as in \cite{brostow2008segmentation, liumulti}.  MPIScene  consists of 4 videos with an average length of 150 frames. The videos show driving scenes recorded from a car. Almost 25\% of the frames are labeled with 5 object classes. For MPIScene, we use the split of 1/2 training and 1/2 test frames as in \cite{wojek2008dynamic}. The SUNY Buffalo dataset consists of 10 videos of diverse scenes with an average length of 80 frames. They are fully annotated with 24 class labels.  As in \cite{jain2013coarse}, we use one-half of the frames for training and the other half for testing.

\textbf{Supervoxels and Descriptors.} All videos are partitioned into supervoxels using the hierarchical graph-based approach of \cite{grundmann2010efficient}, and its software implementation presented in \cite{xu2013flattening}. For CamVid, the video partitioning is based on the 9th level of the generated supervoxel tree, producing on average 3500 supervoxels per video. For, MPIScene, we used the 9th level of the supervoxel hierarchy, producing on average 1000 supervoxels per video. In this paper, we do not consider the computation time of extracting supervoxels.

Each supervoxel has access to algorithms for computing appearance and motion descriptors, as controlled by our inference policy. These descriptors include the following. We use dense trajectories \cite{wang:2011} by tracking a set of densely sampled points in two different scales on a grid. The descriptor extraction finds the tightest cube of a given supervoxel, and uses the dense trajectories to generate HOG, HOF, and MBH (motion boundary histogram) for each track. We also use the color histogram in CIE-Lab color space for each supervoxel. In addition, we also use object detectors as mid-level descriptors of supervoxels to identify the probability of observing a corresponding object class in a supervoxel. Specifically, given a supervoxel, we run a Deep Convolutional Neural Net (DCNN) on all pixels of the supervoxel's first, middle and last frame. Then, the output of the logistic regression layer of DCNN is used as a descriptor of the superpixel. For CamVid, computing HOG, HOF and MBH for the entire video takes 6-8, 15-17, 9-11 minutes, respectively; computing the color histogram for all supervoxels in a video takes 3-5 minutes; and running DCNN takes 0.1 seconds per supervoxel. On average the full descriptor extraction for an entire video in CamVid requires 43-53 minutes. For MPIScene,  computing HOG, HOF and MBH for an entire video takes 10-12, 18-20, 11-12 seconds, respectively; computing the color histogram for all supervoxels in a video takes  2-4 seconds; and running DCNN takes  0.1 seconds per supervoxel. On average the full feature extraction for an entire video in MPIScene requires 42-50 seconds. For SUNY Buffalo-Xiph.org 24-class computing HOG, HOF and MBH for the entire video takes 5-7, 8-10, 5-6 seconds respectively. Computing the color histogram over all supervoxels in a video takes 0.6-1 second. On average, full feature extraction for the entire video from the SUNY Buffalo dataset requires 19-22 seconds.

\textbf{Variations of Our Framework:} We evaluate our framework using different supervoxel selection strategies (Sec.~\ref{sec:budgeted}) and different sets of available descriptors. We consider three variations: 1) \textit{\ourmethod-RndRnk} randomly selects a supervoxel from $\mathcal{C}$, and uses only HOG, HOF, MBH and color histogram; 2) \textit{\ourmethod-NhbRnk} ranks the supervoxels based on the confidence of $H$ classifiers for neighbors in $\mathcal{F}$ (Sec.~\ref{sec:budgeted}), and uses only HOG, HOF, MBH and color histogram; 3) \textit{\ourmethod-Full} is similar to \ourmethod-NhbRnk, but additionally uses DCNN-based descriptor. 

\textbf{Upper-Bound Performance:} For evaluating our upper-bound performance, we compare our time-bounded accuracies to two unbounded CRF models. The first model, referred to as CRF, uses only HOG, HOF, MBH and color histogram descriptors. The second model, referred to as CRF-Full, additionally uses DCNN-based descriptor. 

\textbf{Baselines:} We specify a number of reasonable baseline approaches. This comparison serves to evaluate how our proposed learning of the inference policy affects performance relative to alternative strategies. 1) \textit{Baseline1} randomly selects a sequence of descriptors to be computed for all supervoxels until time runs out; 2) \textit{Baseline2} randomly selects a subset of supervoxels, such that there is time for computing all descriptors for each supervoxel in the subset, and 3) \textit{Baseline3} is aimed to recreate the related approach of \cite{NIPS2013_5142} in our domain, i.e., \textit{Baseline3} learns an inference policy using Q-learning for selecting a sparse set of descriptors that are computed for all supervoxels in the entire video.

\textbf{Implementation:} is done in C++. Darwin library(http://drwn.anu.edu.au/) is used for training the CRF and $H$. The $\alpha$-expansion algorithm \cite{a22001fast} is used for CRF inference. We perform our experiments on an Intel quad core-i7 CPU and 16GB RAM PC. We use Caffe deep learning framework \cite{jia2014caffe} for our DCNN implementation. Fine tuning of the DCNN parameters based on the Alexnet model is done for each dataset. The training set for the objects is obtained as in \cite{liumulti}.

\begin{table*} 
\begin{minipage}[b]{\linewidth} 
  \centering 
    \resizebox{16cm}{!} { 
  \begin{tabular} {|l|l|l|l|l|l|l|l|l|l|l|l|l|l|} 
  \hline 
     B & Method & \begin{turn}{45}Road\end{turn} &    \begin{turn}{45}Bldg\end{turn}&    \begin{turn}{45}Sky\end{turn}&\begin{turn}{45}Tree\end{turn}&    \begin{turn}{45}SWlk\end{turn}&    \begin{turn}{45}Car\end{turn}&    \begin{turn}{45}Pole\end{turn}&\begin{turn}{45}Fence\end{turn}&    \begin{turn}{45}Pdstr\end{turn}&    \begin{turn}{45}Bcyl\end{turn}    &    \begin{turn}{45}Sign\end{turn}& \begin{turn}{45}Avg\end{turn}\\\hline
    \multirow{2}[4]{*}{$\infty$} & CRF   & 90.2  & 74.2  & 95.2  & 79.8  & 69.8  & 75.8  & 10.1  & 29.2  & 59.9  & 35.4  & 50.2  & 60.9 \\
          & CRF-Full & 90.5  & 74.6  & 95.2  & 80.1  & 70.3  & 78.8  & 10.4  & 30.1  & 59.4  & 37.2  & 50.4  & 61.5 \\\hline\hline
    \multirow{6}[12]{*}{10} & Baseline1 & 81.1  & 65.3  & 74.2  & 39.9  & 30.2  & 46.3  & 3.9   & 7.2   & 19.5  & 10.3  & 17.1  & 35.9 \\
          & Baseline2 & 91.9  & 70.9  & 84.4  & 51.6  & 36.2  & 53.2  & 5.8   & 10.4  & 26.2  & 14.1  & 27.2  & 42.9 \\
          & Baseline3 & 89.1  & 71.3  & 83.2  & 47.9  & 33.8  & 50.4  & 6.2   & 11.1  & 25.8  & 13.4  & 24.5  & 41.5 \\
          & \ourmethod-RndRnk & 93.2  & 75.6  & 90.3  & 69.4  & 51.3  & 58.8  & 6.2   & 12.2  & 27.2  & 13.2  & 23.2  & 47.3 \\
          & \ourmethod-NhbRnk & 91.2  & 76.4  & 91.5  & 71.2  & 50.6  & 56.9  & 7.4   & 13.5  & 28.1  & 15.4  & 25.7  & 48.0 \\
          & \ourmethod-Full & 91.9  & 78.9  & 94.2  & 73.4  & 53.8  & 62.4  & 8.1   & 14.1  & 36.6  & 24.5  & 28.8  & 51.5 \\\hline
    \multirow{6}[12]{*}{25} & Baseline1 & 86.9  & 73.8  & 79.8  & 50.7  & 49.1  & 52.2  & 7.2   & 13.6  & 30.2  & 21.5  & 26.4  & 44.7 \\
          & Baseline2 & 89.3  & 75.9  & 88.2  & 68.8  & 40.5  & 60.7  & 7.5   & 15.4  & 38.5  & 25.3  & 29.2  & 49.0 \\
          & Baseline3 & 89.4  & 72.3  & 89.2  & 69.1  & 35.1  & 57.2  & 6.8   & 13.9  & 33.7  & 20.1  & 25.3  & 46.6 \\
          & \ourmethod-RndRnk & 90.6  & 78.9  & 93.4  & 75.1  & 52.9  & 65.5  & 6.7   & 13.5  & 32.6  & 20.7  & 29.1  & 50.8 \\
          & \ourmethod-NhbRnk & 92.9  & 77.7  & 93.2  & 76.6  & 57.4  & 67.5  & 8.1   & 16.7  & 37.4  & 24.5  & 28.9  & 52.8 \\
          & \ourmethod-Full & 92.7  & 77.4  & 96.9  & 79.1  & 63.3  & 73.1  & 9.7   & 20.7  & 44.2  & 29.8  & 36.2  & 56.6 \\\hline
    \multirow{6}[12]{*}{45} & Baseline1 & 88.4  & 72.8  & 92.3  & 78.8  & 68.7  & 76.5  & 10.0  & 24.4  & 57.9  & 35.8  & 48.1  & 59.4 \\
          & Baseline2 & 87.5  & 73.2  & 91.7  & 73.5  & 65.2  & 75.4  & 9.9   & 23.7  & 51.6  & 32.8  & 47.8  & 57.5 \\
          & Baseline3 & 89.6  & 70.9  & 91.2  & 77.5  & 66.7  & 73.9  & 10.4  & 25.8  & 57.8  & 33.9  & 50.5  & 58.9 \\
          & \ourmethod-RndRnk & 89.8  & 66.2  & 92.7  & 77.3  & 64.3  & 72.5  & 9.1   & 24.6  & 53.8  & 32.4  & 47.1  & 57.3 \\
          & \ourmethod-NhbRnk & 90.3  & 72.9  & 89.8  & 76.9  & 64.8  & 76.7  & 9.3   & 27.8  & 56.9  & 34.4  & 46.5  & 58.8 \\
          & \ourmethod-Full & 90.8  & 74.5  & 91.7  & 79.4  & 68.1  & 75.0  & 10.2  & 28.3  & 58.3  & 38.2  & 49.9  & 60.4 \\\hline
  \end{tabular} 
  } 
 \newline 
	\quad 
 \end{minipage} 
 \caption{Per-class and average accuracy on CamVid. We evaluate CamVid for $B \in$\{10, 25, 45 minutes\}. The upper-bound accuracy for \ourmethod-RndRnk and \ourmethod-NhbRnk is the accuracy of the CRF and the upper-bound accuracy for \ourmethod-Full is the accuracy for CRF-Full. The following abbreviations are used: Bldg = Building, SWalk = Side Walk, Pole = Column-Pole, Pdstr = Pedestrian, Bcyl = Bicycle, Sign = Sign Symbol} 
  \label{tab:results} 
\end{table*} 
 
Table~\ref{tab:results} compares per-class and average video labeling accuracy of the aforementioned variations of our framework with those of  time-unconstrained CRF models and baselines, for three different time budgets, on CamVid. As can be seen, for the smallest budget, the accuracy of all baselines is much worse than that of all our framework variations. We observed that Baseline1 and Baseline2 were not able to use HOF and MBH, for this budget, because the cost of computing motion descriptors for the entire video was above the budget.  This demonstrates the importance of our intelligent descriptor selection.  

In Table~\ref{tab:results}, we also see that all variations of our framework are able to continually improve accuracy as the time budget increases.    At the largest time budget of 45 minutes, \ourmethod-Full achieves nearly the same accuracy as the unbounded CRF-Full which in turn requires 50-55min for computation of all descriptors in the entire video. This demonstrates that we are able to maintain a similar level of accuracy of the original method under reduced runtimes, namely for a 5 min time reduction.

Interestingly, the results in Table~\ref{tab:results} show that for lower budgets the variations of our framework give superior accuracy for the dominant class labels relative to unbounded inference.  This is likely due to our explicit capturing of the contextual information from neighboring supervoxels. The results suggest that our policy successfully learned to avoid computing redundant descriptors when the neighbors can provide strong evidence of the label. Thus, the main impact of increasing the budget is to improve accuracy on the non-dominant labels.  
 
Tables \ref{tab:sunny-mpi} show the per-class and average video labeling accuracy on MPI-Scene and SUNY Buffalo-Xiph.org 24-class dataset. 

\begin{table*}

	\begin{minipage}[b]{10cm} 
		\subcaptionbox{Results on MPI-Scene}[\linewidth][c]{% 
			\centering 
			\resizebox{10cm}{!} { 
				\begin{tabular} {|l|l|l|l|l|l|l|l|} 
					\hline 
					B & Method & Bkgd &    Road&    Lane&    Vehicle&    Sky& Avg   \\\hline 
					\multirow{2}[4]{*}{$\infty$} & CRF   & 87.3  & 91.2  & 11.5  & 66.2  & 94.5  & 70.1 \\
					& CRF-Full & 89.8  & 90.4  & 12.1  & 69.8  & 94.9  & 71.4 \\\hline
					\multirow{6}[12]{*}{15} & Baseline1 & 60.3  & 80.4  & 3.6   & 31.4  & 82.4  & 51.6 \\
					& Baseline2 & 63.7  & 81.9  & 3.5   & 32.7  & 83.7  & 53.1 \\
					& Baseline3 & 65.9  & 81.8  & 3.9   & 31.9  & 83.5  & 53.4 \\
					& \ourmethod-RndRnk & 70.3  & 82.9  & 7.3   & 39.2  & 84.1  & 56.8 \\
					& \ourmethod-NhbRnk & 73.5  & 85.2  & 8.9   & 42.4  & 86.7  & 59.3 \\
					& \ourmethod-Full & 74.6  & 85.6  & 9.1   & 44.8  & 87.2  & 60.3 \\\hline
					\multirow{6}[12]{*}{30} & Baseline1 & 74.8  & 84.2  & 6.9   & 36.4  & 85.5  & 57.6 \\
					& Baseline2 & 75.9  & 87.3  & 6.5   & 39.3  & 84.9  & 58.8 \\
					& Baseline3 & 75.0  & 85.7  & 7.2   & 38.9  & 83.5  & 58.1 \\
					& \ourmethod-RndRnk & 80.1  & 83.4  & 9.4   & 49.7  & 88.7  & 62.3 \\
					& \ourmethod-NhbRnk & 83.6  & 86.8  & 9.8   & 52.9  & 89.3  & 64.5 \\
					& \ourmethod-Full & 84.1  & 88.4  & 10.1  & 54.9  & 89.4  & 65.4 \\\hline
					\multirow{6}[12]{*}{45} & Baseline1 & 88.3  & 91.4  & 10.9  & 66.4  & 93.2  & 70.0 \\
					& Baseline2 & 86.8  & 90.1  & 9.3   & 63.9  & 94.1  & 68.8 \\
					& Baseline3 & 89.2  & 91.3  & 11.1  & 66.6  & 94.7  & 70.6 \\
					& \ourmethod-RndRnk & 86.8  & 90.7  & 10.1  & 64.8  & 93.9  & 69.3 \\
					& \ourmethod-NhbRnk & 88.9  & 91.2  & 10.4  & 65.6  & 94.7  & 70.2 \\
					& \ourmethod-Full & 89.8  & 91.5  & 11.1  & 70.9  & 94.9  & 71.6 \\\hline
				\end{tabular} 
			} 
			\newline 
		}\quad 
	\end{minipage} 
		\hspace{20pt} 
		\begin{minipage}[b]{5cm} 
			\subcaptionbox{Results on SUNY Buffalo}[\linewidth][c]{% 
				\centering 
				\resizebox{5cm}{!} { 
	\begin{tabular} {|l|l|l|}
		\hline
		
		B & Method & Acc   \\\hline
		$\infty$    & CRF & 52.1 \\\hline\hline
		\multirow{4}[12]{*}{10} & Baseline1 & 32.2 \\
		& Baseline3 & 32.8 \\
		& \ourmethod-RndRnk & 38.7 \\
		& \ourmethod-NhbRnk & 39.6 \\\hline
		\multirow{4}[12]{*}{15} & Baseline1 & 37.4 \\
		& Baseline3 & 37.9 \\
		& \ourmethod-RndRnk & 44.7 \\
		& \ourmethod-NhbRnk & 46.6 \\\hline
		\multirow{4}[12]{*}{20} & Baseline1 & 49.3 \\
		& Baseline3 & 49.8 \\
		& \ourmethod-RndRnk & 48.7 \\
		& \ourmethod-NhbRnk & 49.5 \\\hline
	\end{tabular}
				} 
				\newline 
			}\quad 
		\end{minipage}
	\caption{Per-class and average accuracy on MPI-Scene (Left) and SUNY Buffalo (Right). We evaluate MPI-Scene for $B \in$\{15, 30, 45 seconds\} and SUNY Buffalo for $B \in$\{10sec, 15sec, 20sec\}. The upper-bound accuracy for \ourmethod-RndRnk and \ourmethod-NhbRnk is the accuracy of the CRF and the upper-bound accuracy for \ourmethod-Full is the accuracy for CRF-Full. The following abbreviation is used:  Bkgd = Background} 
	\label{tab:sunny-mpi} 
\end{table*}

\begin{figure}
 \begin{minipage}[b]{7cm}
  \centering
    \resizebox{7cm}{!} {
    \includegraphics[width=0.7\linewidth]{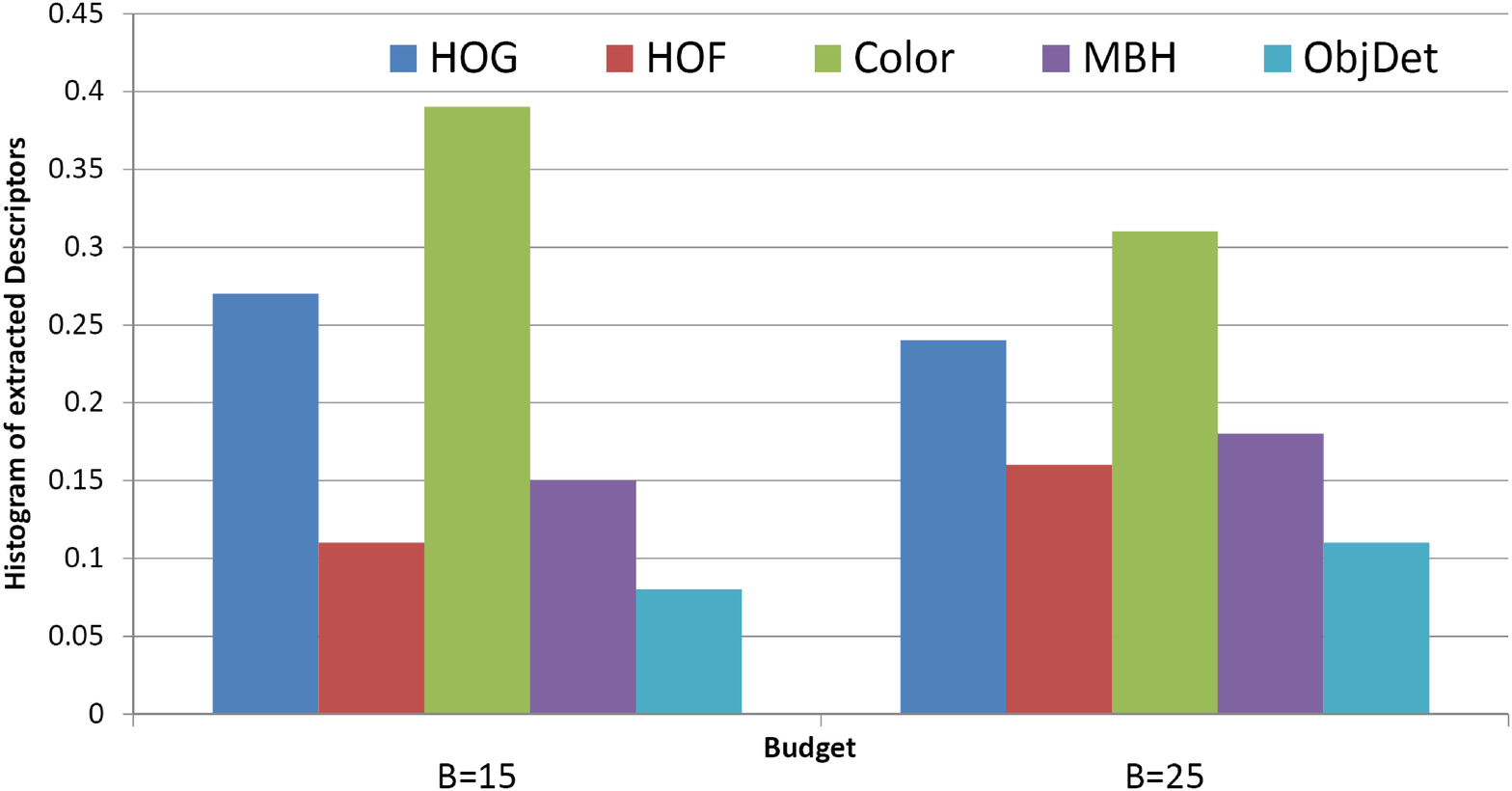}
    }
  \end{minipage}
 \begin{minipage}[b]{7cm}
  \centering
    \resizebox{7cm}{!} {
    \includegraphics[width=0.7\linewidth]{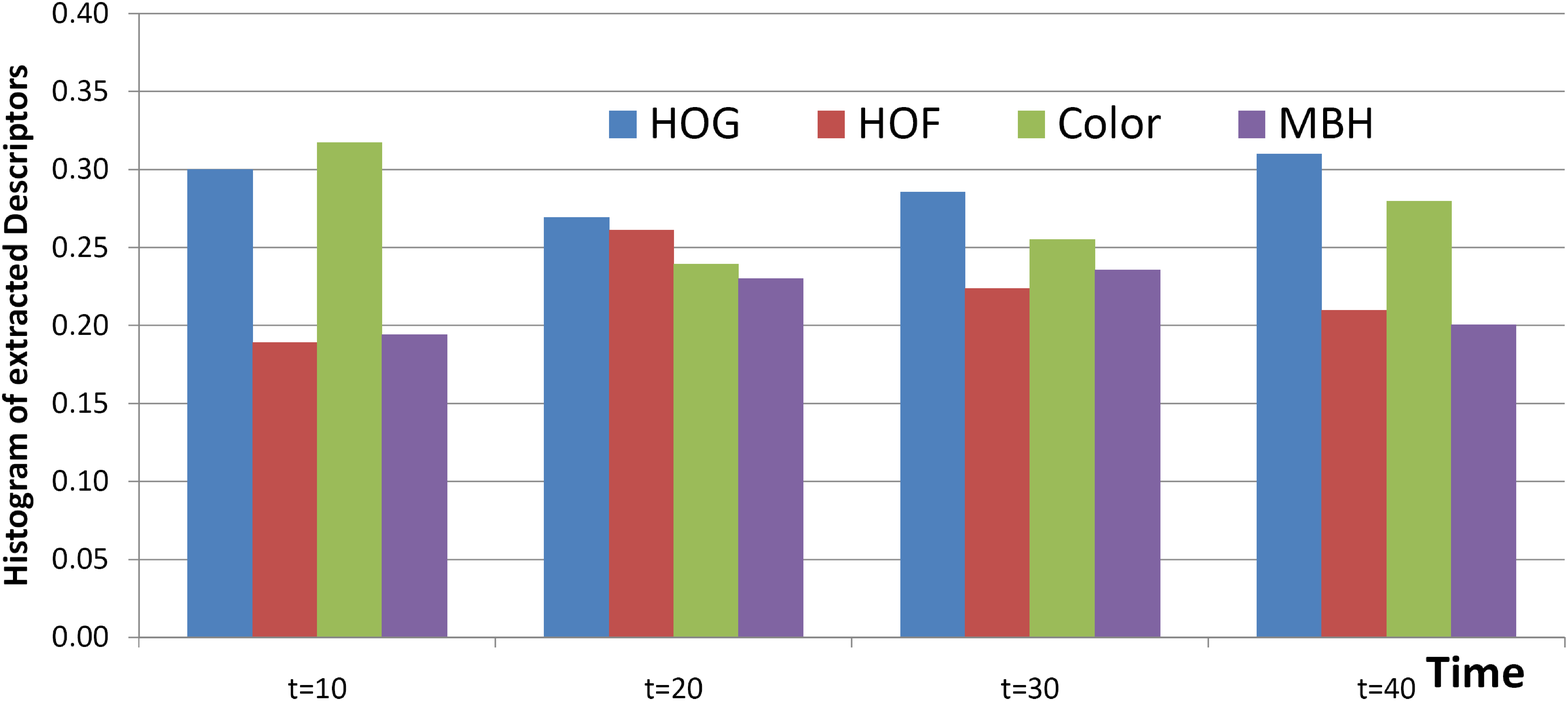}
    }
  \end{minipage}
  \caption{(top) Histogram of descriptors selected by  \textit{\ourmethod-Full}  for different budgets on CamVid. (bottom) Histogram of descriptor selection by  \textit{\ourmethod-NhbRnk} in time on CamVid, constrained by ${B} = 45min$.} \label{fig:plot}
\end{figure}

Fig.~\ref{fig:plot}(top) how decisions of our policy, learned in \textit{\ourmethod-Full}, differ across various budgets  for CamVid. Specifically, the figure shows the histogram of certain types of descriptors selected for different budgets. We see that HOF and MBH are seldom used for small budgets, as expected, since they incur higher costs. Also, the color histogram descriptor is more frequently selected when we are given small budgets, as expected, since they incur small costs.  

Fig.~\ref{fig:plot}(bottom) shows how the distribution of descriptor selection changes in time during our budgeted inference by \textit{\ourmethod-NhbRnk} for CamVid. Specifically, the figure shows the histogram of descriptor selections made by the policy at various moments in time of the inference process. We see that the distribution changes as the inference time increases until the budget is reached. For example, initially, the distribution is highly skewed toward selections of the cheap color descriptor but then becomes more uniform. This suggests that the policy successfully learned to select low-cost descriptors initially for facilitating later policy decisions.  
%%%%%%%%%%%%%%%%%
\section{Conclusion}

We formulated the new problem of budgeted semantic video segmentation, where the pixels of a video must be semantically labeled under a time budget. We presented a budgeted inference framework for this problem that intelligently selects supervoxel descriptors to run, which are then used for CRF inference. Since descriptor computation often dominates the cost of CRF inference, our framework can provide substantial time savings in a principled manner. We formulated the inference policy for selecting among descriptors to run for each supervoxel in a video. We introduced a principled algorithm for learning such policies based on labeled training videos by formulating budgeted inference in the framework of an MDP. Our experiments show that we are able to learn policies for budgeted inference that significantly improve on the accuracies of several baselines. The results also demonstrate that we can optimally adapt a method, design to operate with no time bound, to varying time budgets, such that it yields satisfactory performance for any budget, and maintains an accuracy as close as possible to its performance for no time bound. 

{\small
\bibliographystyle{ieee}
\bibliography{library}
}

\end{document}